\documentclass[10pt,twocolumn,letterpaper]{article}

\usepackage{times}
\usepackage{epsfig}
\usepackage{graphicx}
\usepackage{amsmath}
\usepackage{amssymb}
\usepackage{graphicx}

\usepackage[pagebackref=true,breaklinks=true,letterpaper=true,colorlinks,bookmarks=false]{hyperref}

\begin{document}

\title{3D Reconstruction of Simple Objects from A Single View Silhouette Image}

\author{Xinhan Di\\Trinity College Ireland\\ dixi@tcd.ie \and Pengqian Yu\\National University of Singapore\\ yupengqian@u.nus.edu   }
\date{Nov 2016}

\maketitle

\begin{figure*}[t]
\centering
\includegraphics[width=1.0\linewidth,natwidth=610,natheight=642]{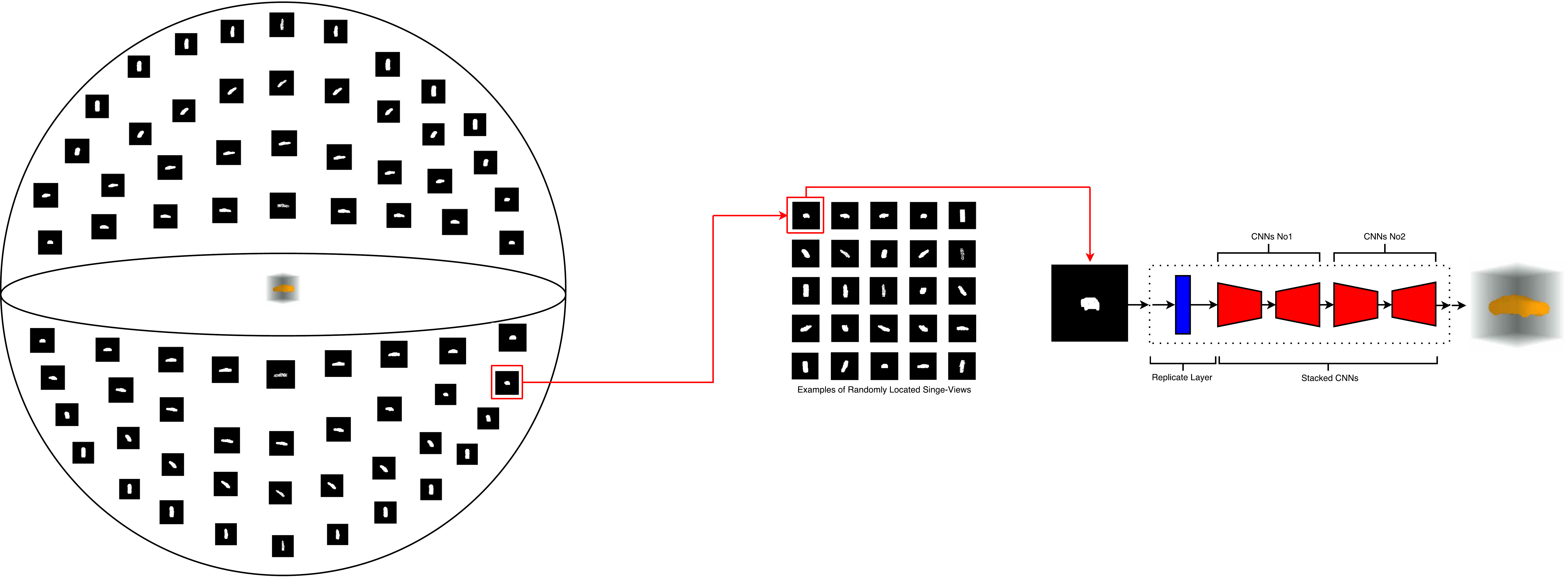}
\caption{Category Specific 3D Reconstruction from A Single-view Silhouette Image. The Left part represents single-view silhouette images located randomly on a sphere around a object. The right part represents a proposed framework to reconstruct a 3D shape from a single-view silhouette image.}
\label{fig:1}
\end{figure*}


\begin{abstract}
While recent deep neural networks have achieved promising results for 3D reconstruction from a single-view image, these rely on the availability of RGB textures in images and extra information as supervision. In this work, we propose novel stacked hierarchical networks and an end to end training strategy to tackle a more challenging task for the first time, 3D reconstruction from a single-view 2D silhouette image. We demonstrate that our model is able to conduct 3D reconstruction from a single-view silhouette image both qualitatively and quantitatively. Evaluation is performed using Shapenet for the single-view reconstruction and results are presented in comparison with a single network, to highlight the improvements obtained with the proposed stacked networks and the end to end training strategy. Furthermore, 3D reconstruction in forms of IoU is compared with the state of art 3D reconstruction from a single-view RGB image, and the proposed model achieves higher IoU than the state of art of reconstruction from a single view RGB image.
\end{abstract}

\section{Introduction}

3D reconstruction techniques rebuild 3D world from 2D information and contribution has been consistently made since the latest two decades. Theoretically, framework are built in this area such as space carving \cite{kutulakos2000theory}, SfM \cite{snavely2010bundler}, SLAM \cite{davison2007monoslam} and CMVS/PMVS \cite{furukawa2010accurate}. Practically, implementation to reconstruct the 3D world is developed such as VisualSfM \cite{wu2013towards} and CMPMVS \cite{jancosek2011multi}.\par
Among a variety of the techniques aiming at 3D reconstruction, 3D reconstruction from a single view is challenging. In order to overcome this challenge, shape priors are applied in order to compensate for the problem of ill-posedness. For example, smoothness in the definition of consistency of surfaces is learned and applied for the reconstruction of curved surfaces \cite{oswald2009non,prados2005shape,Prasad09,toppe2010image,zhang2002single,hoiem2005automatic,saxena2009make3d}. Geometric relation is developed as another kind of prior to dissolve ambiguities in the reconstruction process \cite{lowe1987three,criminisi2000single,delage2007automatic,han2003bayesian,hoiem2005automatic}. High lever priors are also used to overcome single-view reconstruction. Semantic relation priors which infer structure information of objects are applied \cite{han2003bayesian,koutsourakis2009single,delage2007automatic,hoiem2005automatic}. Also, the learned shape priors are used for the reconstruction of facades \cite{koutsourakis2009single}, human bodies \cite{chen2011single}, and category specific objects \cite{vicente2014reconstructing,kar2015category} through learning from defined consistency rules.\par
Thanks to the development of the deep learning techniques, 3D reconstruction from a single view reduces the dependence on classical priors through employing a trained reconstruction model. A 3D recurrent reconstruction neural network(3D-R2N2) is developed to learn a mapping from images of objects to the 3D shapes and is presented as rebuilding 3D shape from a single image \cite{choy20163d}, 3D Interpreter network is designed for single-view reconstruction with application of two pre-trained models and a projection layer during training a deep learning framework \cite{wu2016single}, the prediction of both an RGB image and a depth map of the object are made through a convolutional network \cite{tatarchenko2016multi}, another joint prediction of a depth map and intrinsic images from single-image input is conducted through training a joint convolutional neural field(JCNF) model \cite{kim2016unified}. Stereo pair images are used to train an auto-encoder without requiring a pre-training stage for single-view depth prediction \cite{garg2016unsupervised}. Volumetric 3D object reconstruction is achieved from the designed network with extra input of the pre-trained projection transformation code \cite{yan2016learning}.\par
In this paper, we propose a stacked network for 3D reconstruction from a single-view silhouette image that is demonstrated to be able to reconstruct category-specific objects from a randomly located single-view silhouette image. The overall architecture of the architecture is illustrated in Fig\ref{eq:1}. The architecture consists of two major components:(1) the replicate layer, which works to produce a single-view 2D segmentation to 3D volume coarse segmentation through simple replication operation. (2) stacked reconstruction networks, which uses the 3D volume segmentation to reconstruct a 3D shape. The proposed network architecture is trained end to end through following a proposed training strategy for stacked hierarchical network without a pre-training stage or separate training for each partial network. The proposed training strategy works as controlling the output of each network during training process of the staked hierarchical networks through gradient calculation. In the stacked hierarchical network architecture, it follows a unique criterion to minimize the error for the final network which is different from combined criterion for each network. As illustrated in Fig\ref{eq:1}, the proposed architecture achieves good performance for 3D category specific reconstruction from a single-view silhouette image, and the silhouette image is randomly located around a sphere, and previous network framework for 3D reconstruction from single view is not demonstrated to deal with this large view variation .\par 
The main contributions of our work are three-fold. Firstly, we proposed a stacked hierarchical network for 3D reconstruction from a single view and its end-to-end training strategy. And we demonstrate that the design of the stacked networks architecture and its training strategy works much better than a single network. Secondly, we demonstrate that category specific 3D reconstruction from a single-view silhouette image without RGB textures or other extra input is conducted through our proposed model. Thirdly, we demonstrate that for 3D reconstruction from a single-view silhouette image, single views are randomly and widely located on a sphere surface while previous network frameworks work for small view variation, such as single view images located on a circle.

\section{Related Work}
Network is employed as a data-driven method to provide useful priors for 3D reconstruction from a single view image recently. WarpNet is exploited to align an object in one image with a different object in another which allows single-view reconstructions with quality \cite{kanazawa2016warpnet}. Virtual view networks(VNN) \cite{carreira2015virtual} are built to produce smooth rotations through the class object collection and points matching for point cloud reconstruction from a single image. The view prior of a single image is estimated through exploiting a deep CNN architecture with a high learning capacity \cite{su2015render}. Two deep network stacks are employed to produce a depth prior for a single image without the need for superpixelation \cite{NIPS2014_5539}.\par
Furthermore, deep learning networks are applied to build frameworks for 3D reconstructions from a single-view image. Among many frameworks that are applied, the most relevant works are learned NRSfM model with defined consistency and smoothness criterion \cite{kar2015category}, 3D-R2N2 network with adding 3D convolutional LSTM inside \cite{choy20163d}, 3D-INN framework with usage of 2 pre-trained models and a projection layer \cite{wu2016single} , volumetric reconstruction of objects with pre-trained projective transformation code \cite{yan2016learning} and depth prediction from CNN neural networks \cite{tatarchenko2016multi,kim2016unified,garg2016unsupervised}. A single network is built for 3D reconstruction from multiple-view silhouette images with multiple-view information \cite{DiECCV2016}. \par
Compared with the above frameworks, our framework is considered as advantages than the exploited networks for 3D reconstruction from a single view. Firstly, despite that a single-view RGB image \cite{choy20163d,wu2016single,yan2016learning,tatarchenko2016multi} is used as input, our framework releases RGB dependence to a single-view silhouette image. Secondly, the tested single views \cite{yan2016learning,choy20163d,wu2016single,tatarchenko2016multi} are not largely located in space, but our framework works to reconstruct a 3D shape from a single view randomly and widely located on a sphere around the object, while the above frameworks achieves 3D reconstruction from a number of single views or arbitrary views around circles. Thirdly, we reduce the input of single-view silhouette image without key points used in \cite{kar2015category}. Finally, we build a stacked hierarchical networks and training the network end to end despite of separate training of each network or usage of pre-trained codes \cite{yan2016learning,wu2016single}.

\section{Stacked Hierarchical End-to-end Training Networks}

A common choice of completing complex task through networks is a design of stack networks. And train each single network in multiple stages that separate training is conducted for training each network. However, this is not strictly trained end to end. However, we propose a novel design and end to end training strategy for stacked networks and employ it for the challenging task of 3D reconstruction from a single-view silhouette image.

\subsection{Networks Architecture and Training Strategy}
Stacked hierarchical networks are built for training a model to complete a complex task.

\begin{figure}
\centering
\includegraphics[width=0.8\linewidth,natwidth=305,natheight=321]{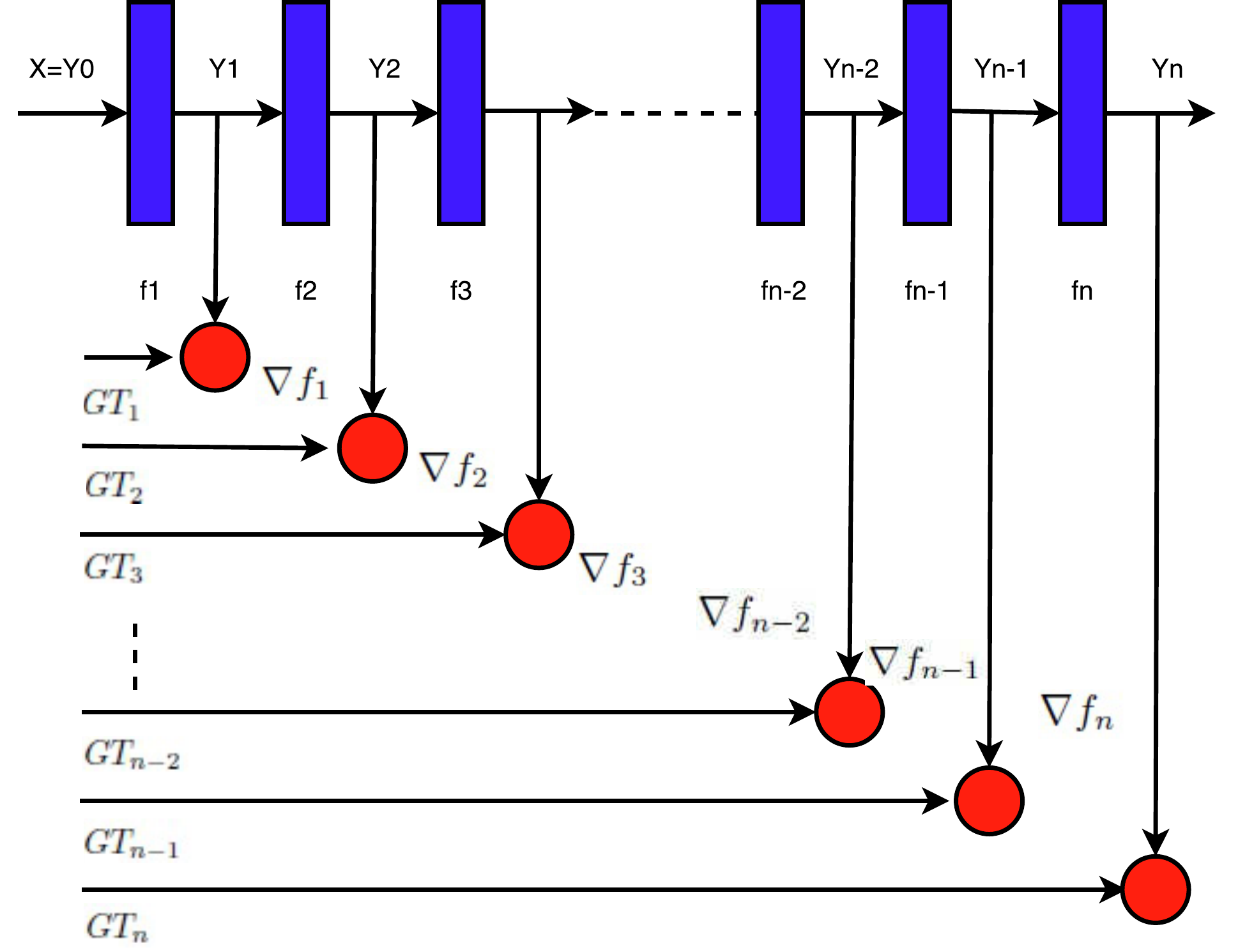}
\caption{Stacked hierarchical networks.}
\label{fig:2}
\end{figure}

\paragraph{Definition 3.1} Let \(Y=F(X)\) be stacked hierarchical networks where \(X\) is the input, \(Y\) is the output of the stacked networks, \(F\) is the overall function of the stack networks, \(n\) is the number of the single networks that this stacked networks are comprised of. Consider \(F\) can be split as n stacked functions \(f_{k},k=1...n\) and \(f_{k}\) is the overall function of \(kth\) sub-network in the stacked hierarchical networks. Also consider \(Y_{k}=f_{k}(Y_{k}-1),k=1...n\), \(Y_{k}\) is the output of the \(kth\) sub-network. And  \(Y_{0} = X\), \(Y_{n} = Y\).
\paragraph{Strategy 3.1} Consider the stacked hierarchical networks \(Y=F(X)\), in order to guarantee an effective end-to-end training for stacked hierarchical networks combined with a number of single networks. Both a multiple gradient decent learning strategy and an unique error criterion for the final single network of the stacked networks are employed. This strategy is employed in order to find a global solution for stacked networks to overcome the difficulty of searching global optimization from end-to-end training.  

Consider multiple gradient descent learning strategy as the Eq.(\ref{eq:1}) and Eq.(\ref{eq:2}).

\begin{equation} \label{eq:1}
\nabla F(X)=\sum_{k=1}^{n}\lambda_{k}\nabla f_{k}(Y_{k-1})
\end{equation}

\begin{equation} \label{eq:2}
\nabla f_{k}(Y_{k-1})=|f_{k}(Y_{k-1})-GT_{k}|^{\frac{1}{2}} 
\end{equation}

where \(\nabla F(X)\) is the multiple gradient for the stacked networks during training, \(\nabla f_{k}\) is the gradient for the \(kth\) single network. The parameter \(\lambda_{k}\) follows \(\sum_{k=1}^{n}\lambda_{k}=1\). \(GT_{k}\) is the ground truth for the \(kth\) single network.

Consider learning the unique error criterion as the Eq.(\ref{eq:3}).

\begin{equation} \label{eq:3}
\Theta^{*}=\arg \underset{\Theta}{\min}|F^(X)-GT_{n}|^{\frac{1}{2}},\Theta={\theta_{1},...,\theta_{n}}
\end{equation}

It works as learning the global optimization parameters set \(\Theta^{*} = {\theta_{1}^{*},...,\theta_{n}^{*}}\) for each single network in the stacked hierarchical networks. The criterion only focus on the minimization of the error of the final single network in the stacked networks despite of a combination criterion for the minimization of error from all single networks. Both the stacked hierarchical network architecture and it's end-to-end training strategy are illustrated in Fig\ref{fig:2}.  

\begin{figure}
\centering
\includegraphics[width=0.8\linewidth,natwidth=305,natheight=321]{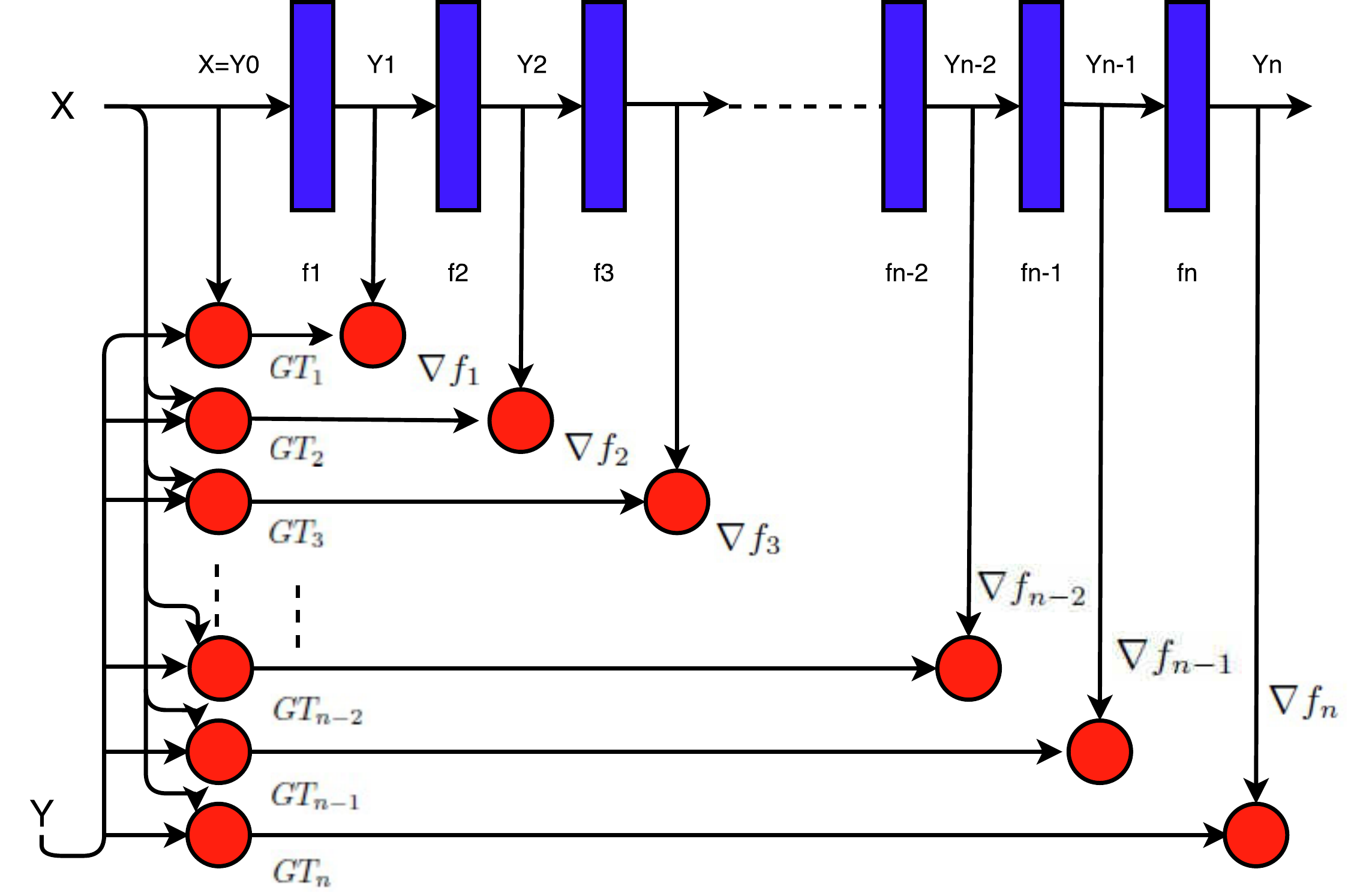}
\caption{Stacked hierarchical networks for the proposed application.}
\label{fig:3}
\end{figure}

\paragraph{Application 3.1} Consider a single network \(Y=f(X)\), where \(X\) is the input, \(Y\) is the output, \(GT\) is the ground truth, \(f\) is the single network overall function, \(S\) is the architecture of the single network. Practically, this network is able to achieve the error boundary \(E^{*} = |y-y^{*}|^{\frac{1}{p}},p\geqq1\). In the real world situation, in order to increase the learning ability of the single network, that is to reduce the error boundary from \(E^{*}\) to \( E^{'},E^{'} < E^{*}\), stacked networks are designed as \(Y=F(X)\), where \(S_{k+1} = S_{k}, k=1,..,n-1\), \(GT_{k} = \eta_{k}^{1} \times X + \eta_{k}^{2} \times GT\),\(\sum_{i=1}^{2}\eta_{k}^{i} = 1\), following the proposed definition and strategy and \(k\) is the \(kth\) single network, \(k\in\{1,2,...,n\}\). The deep architecture for this application is illustrated in Fig\ref{fig:3}.

\subsection{Application for Category Specific 3D Reconstruction from A Single-view silhouette image}
In order to conduct 3D reconstruction from a single-view silhouette image, stacked hierarchical networks are proposed. Following the proposed application and definition above, the stacked networks are combined of two single networks. Both of the single network share the same architecture. Each one contains four 3D convolution layers and four 3D deconvolution layers. The size of the filters for convolution layers are \(4\times 4\times 4,2\times 2\times 2,2\times 2\times 2,2\times 2\times 2\). The size of filters for deconvolution layers are \(2\times 2\times 2,2\times 2\times 2,2\times 2\times 2,4\times 4\times 4\). The parameter of stride for all layers is \(2\). The numbers of filters for the four convolution layers are \(64,256,512,1024\). The numbers of filters for the four deconvolution layers are \(1024,512,256,64\). 

\begin{figure}
\centering
\includegraphics[width=0.8\linewidth,natwidth=305,natheight=321]{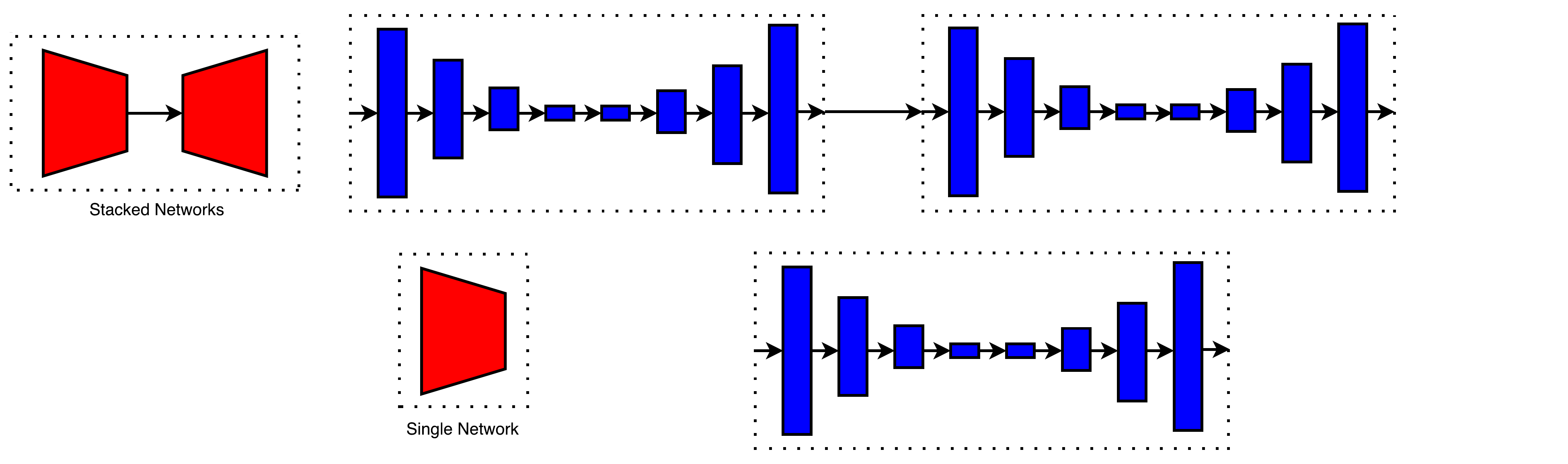}
\caption{Two network architectures. The top are the concise architecture(top left) of stacked hierarchical networks and the detailed architecture(top right). The bottom are the concise architecture(bottom left) of the single network \cite{DiECCV2016} and the details architecture(bottom left), the replicate layer is not represented.}
\label{fig:4}
\end{figure}

Also, before the stacked networks, a replication layer is added to work as replication of single-view 2D segmentation for producing 3D segmentation. In the task of 3D reconstruction from a single-view silhouette image, the single-view silhouette image input is \(50 \times 50 \), and the replication layer simply works to replicate the 2D segmentation \(50\) times to produce a 3D segmentation in the 3D grid space. While voxels in the space gets binary value, 1 represents that the voxel is in the segmentation, 0 represents that the voxel is not in the segmentation. The architecture of the stacked deep architecture is illustrated in Fig\ref{fig:4}.

Furthermore, in order to train the proposed deep architecture from end to end. The proposed training strategy is applied. Here, following the proposed training strategy in Eq.(\ref{eq:4}), Eq.(\ref{eq:5}), Eq.(\ref{eq:6}) and Eq.(\ref{eq:7}).

\begin{equation} \label{eq:4}
S_{1} = S_{2},f_{1} = f_{2} = f
\end{equation}

\begin{equation} \label{eq:5}
\nabla f_{1} = |Y_{1} - 0.5 \times X - 0.5 \times GT|^{\frac{1}{2}}
\end{equation}

\begin{equation} \label{eq:6}
\nabla f_{2} = |Y_{2} - GT|^{\frac{1}{2}}
\end{equation}

\begin{equation} \label{eq:7}
\Theta^{*}=\arg \underset{\Theta}{\min}|F(X)-GT|^{\frac{1}{2}},\Theta=\{\theta_{1},\theta_{2}\}
\end{equation}

\section{Evaluation}
Experiments for 3D reconstruction from a single-view silhouette image are conducted to evaluate the performance of the proposed stacked networks and the end-to-end training strategy. To demonstrate the proposed model's ability for class-specific reconstruction from a single-view silhouette image, experiments are conducted for two object categories including the car and the plane. We also conduct 3D reconstruction experiments for both the staked networks and the single network \cite{DiECCV2016} Fig\ref{fig:4}, and demonstrate the improvement of the proposed stacked network both qualitatively and quantitatively. Furthermore, results are calculated in voxel IoU in order to make comparison with the state of art single-view reconstruction work. 

\subsection{Dataset}
ShapeNet is a richly-annotated, large-scale repository of shapes represented by 3D CAD models of objects. It contains more than 3,000,000 models, 220,000 models out of which are classified into 3135 categories. We take use of 2 categories of the ShapeNet, planes, and cars for the evaluation of the our reconstruction framework. For each CAD model of the training and testing dataset, we project a CAD model from a large set of single views to get single-view silhouette images of the object.

\subsection{Training}
\begin{figure}
\centering
\includegraphics[width=0.3\linewidth,natwidth=305,natheight=321]{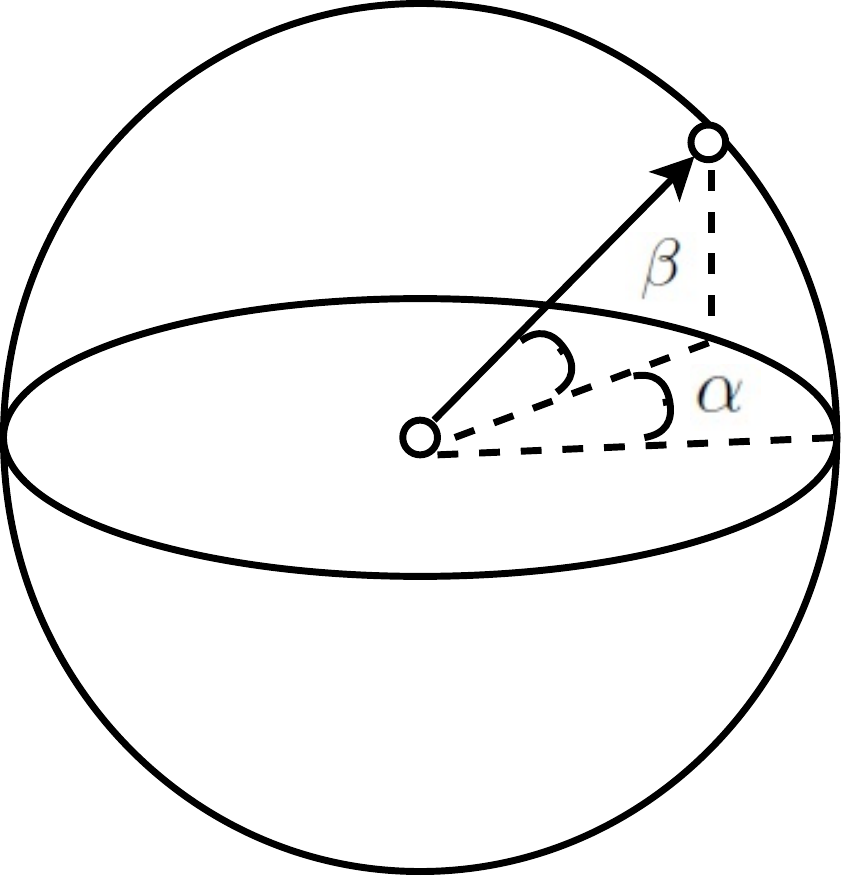}
\caption{A widely located single view on a sphere.}
\label{fig:5}
\end{figure}

We evaluate the proposed single-view reconstruction framework for 2 object categories including cars and planes. For each object category, we randomly pick 100 different CAD models. We train 78 CAD models separately for each object category. As presented, for each CAD model, 180 single-view silhouette images are produced through projection from views widely located on a sphere around the model. Therefore, we train \(78\times180\) single-view silhouette images for each object category. Similarly, for the test, we test 22 CAD models and pick up 180 single-view silhouettes images for each model. Therefore, \(22\times180\) single-view silhouette images are test for each object category. In order to measure the quality of the reconstruction shape quantitatively, we measure the voxel overlap between the reconstructed shape and the ground truth shape(voxel IoU). Also, to represent the accuracy improvement of our stacked networks compared with the single network \cite{DiECCV2016}, we evaluate rebuilt shapes produced from the two networks for comparison. As illustrated in Fig\ref{fig:5}, for the reconstruction experiments of cars and planes, we follow the Eq.(\ref{eq:8}) 

\begin{equation} \label{eq:8}
\alpha_{1}=10l_{1},\beta_{2}=20l_{2},l_{1}\in\{1,2,...,18\},l_{2}\in\{1,2,...,18\}
\end{equation}

 to get 180 single views on a sphere around each object.  And 2D segmentation of the object is projected from each single view. Fig\ref{fig:1} demonstrates an example for the choice of single views picked for a car. 

\subsection{Evaluation on Sphere-surface Located Single-view Silhouette Images (Experiment1)}
For this test, we test 22 CAD models for each category and 180 single-view silhouette images are rendered for the test of each model. A single view for each single-view silhouette image is picked following the same rule in the training. That is to say taht the views picked for test is the same views in the training. These views are widely located on the sphere around of the object. Therefore, there are totally \(22\times180\) single-view silhouette images for test for each object category.

\begin{figure}
\centering
\includegraphics[width=1.0\linewidth,natwidth=305,natheight=321]{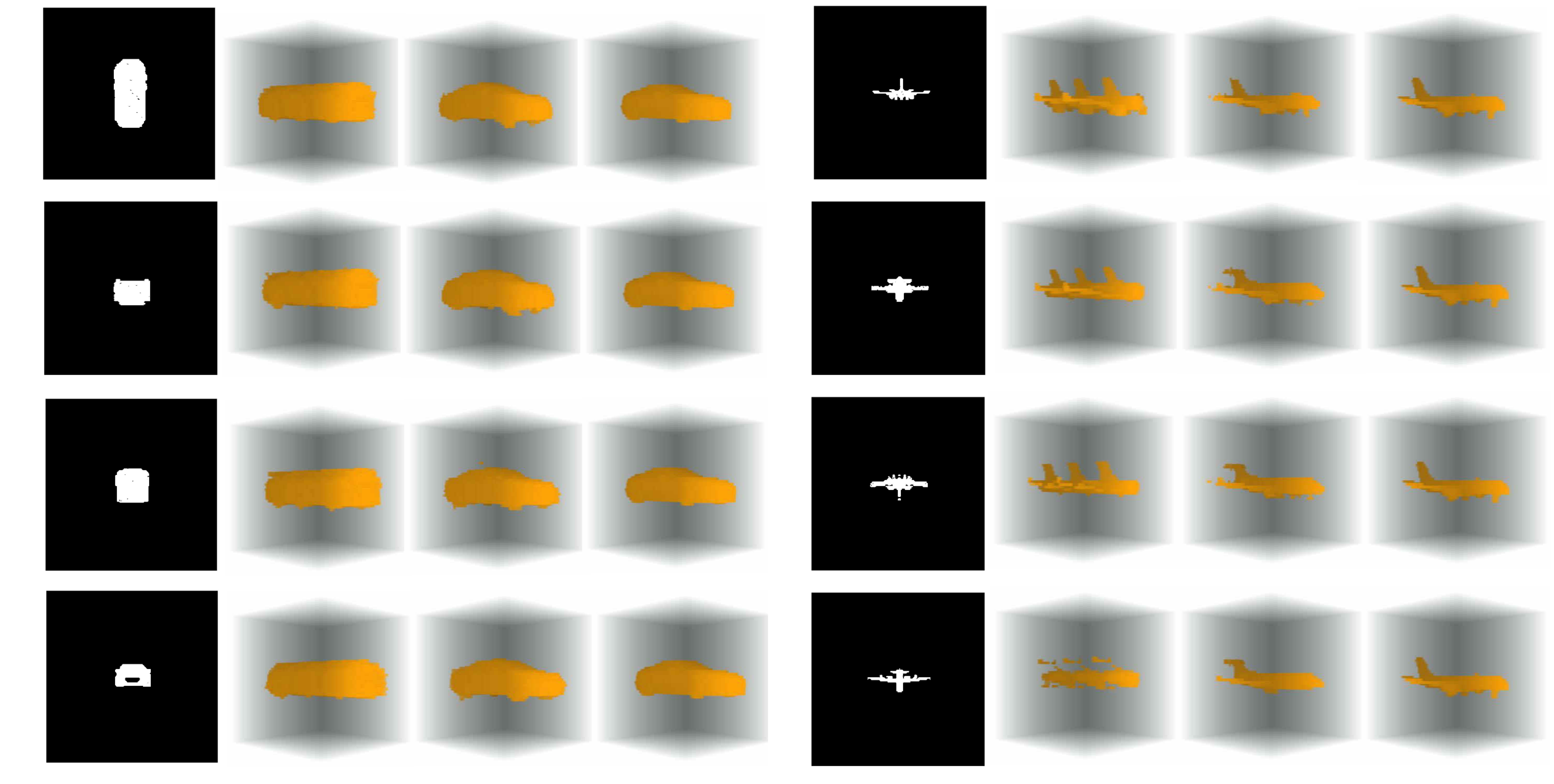}
\caption{Reconstruction visualization from a hard single-view silhouette image(Experiment1). The left part represents visualization for cars and the right part represents visualization for planes．　For each part, the first column represents picked hard single-view silhouette image of an object. The second column represents the rebuilt 3D shape reconstructed from a single network given the corresponding silhouette images. The third column represents the rebuilt 3D shape reconstructed from the proposed stacked networks. The fourth column represents the ground truth shape.}
\label{fig:6}
\end{figure}

As illustrated in Fig\ref{fig:6}, for 3D reconstruction of cars from a single-view silhouette image, and only a small part of 3D shape is available from a single view, such as the bottom view, the top view, the front view and the back view, it's challenging to reconstruct a good quality of 3D shape from a 2D silhouette image where most part of the 3D shape are unavailable(hard single-view silhouette images). It's reasonable that the single network is only able to reconstruct a very coarse 3D shape in this case. However, for the proposed stacked network and its end-to-end training strategy, a good quality 3D shape is reconstructed even most part of the shape is unavailable from the single-view silhouette images. We test 440 hard single-view silhouette images totally for the reconstruction of cars from a single-view silhouette image.

Similarly, as illustrated in Fig\ref{fig:6}, for 3D reconstruction of planes from a single-view silhouette image, the single network is able to rebuild a 3D shape but the shape is not reasonable as the silhouette images from these single views do not contain necessary information, such as where to put the airfoil or the tail, what dose the head look like when seen from the back. However, our stacked networks are able to reconstruct good 3D shape for this case, fix the tail airfoil at the right place and reconstruct a good head even these is no information from these views. Also, the number of hard single-view silhouette images of planes for this test is 440.

\begin{figure}
\centering
\includegraphics[width=1.0\linewidth,natwidth=305,natheight=321]{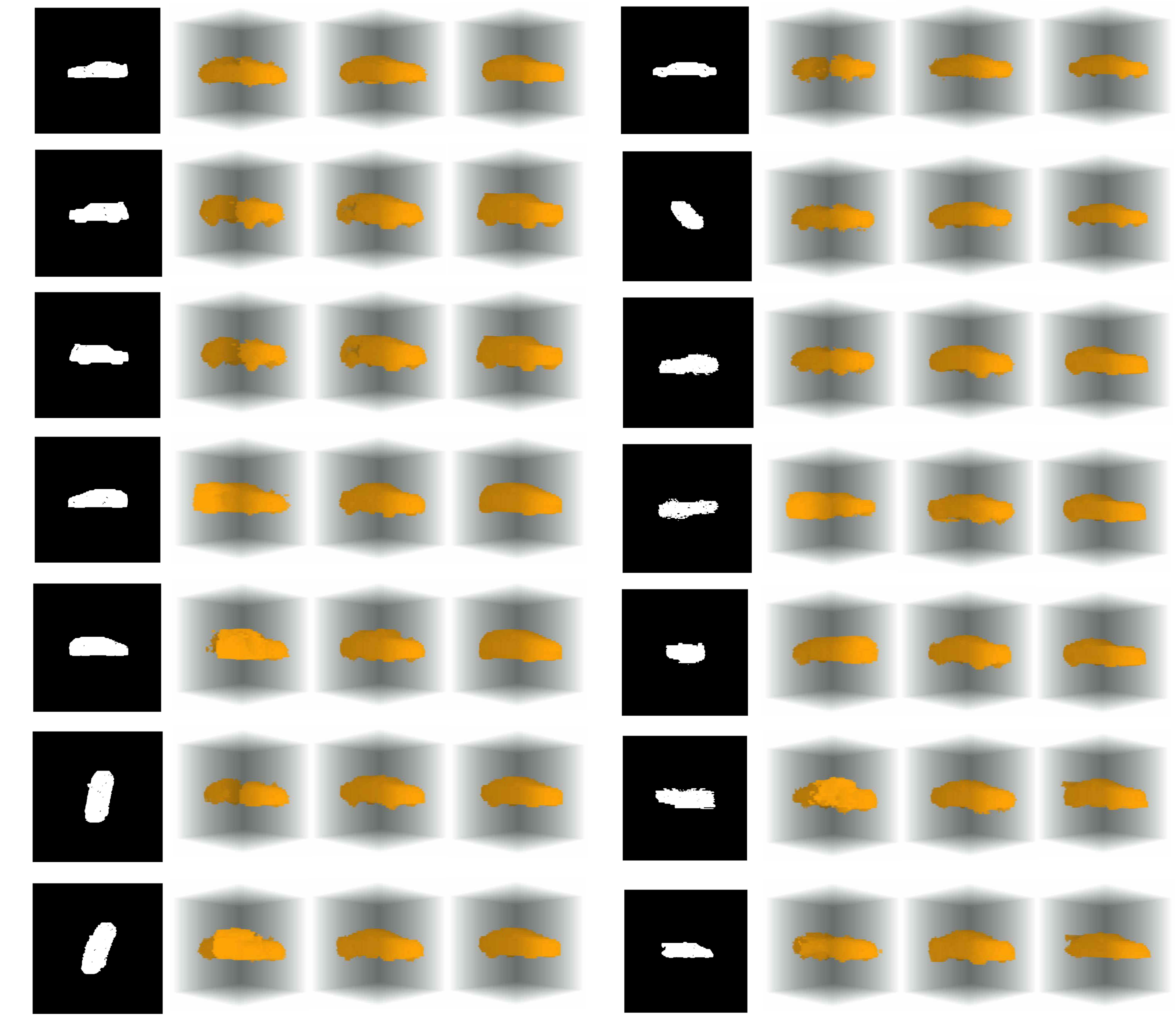}
\caption{Reconstruction visualization from a easy single-view silhouette image for cars(Experiment1). The first column represents picked easy single-view silhouette images of an object. The second column represents the rebuilt 3D shape reconstructed from a single network given the corresponding silhouette images. The third column represents the rebuilt 3D shape reconstructed from the proposed networks. The fourth column represents the ground truth shape.}
\label{fig:7}
\end{figure}

Also, as illustrated in Fig\ref{fig:7}, for 3D reconstruction of cars from a single-view silhouette image, and most part of the shape is available in each silhouette image from a single view(easy single-view silhouette image), such as the side views, the left views and the right views, the single network is able to build a reasonable shape but these shape have obvious drawbacks such as big holes, fractures, partial shape damage, partial coarse shape. However, the proposed stacked networks and its end to end training strategy help the reconstruction to overcome these drawbacks, and the 3D reconstructed shapes are very close to the ground truth shapes. We test 3520 easy single-view silhouette images totally for the reconstruction of cars from a single-view silhouette image.

\begin{figure}
\centering
\includegraphics[width=1.0\linewidth,natwidth=305,natheight=321]{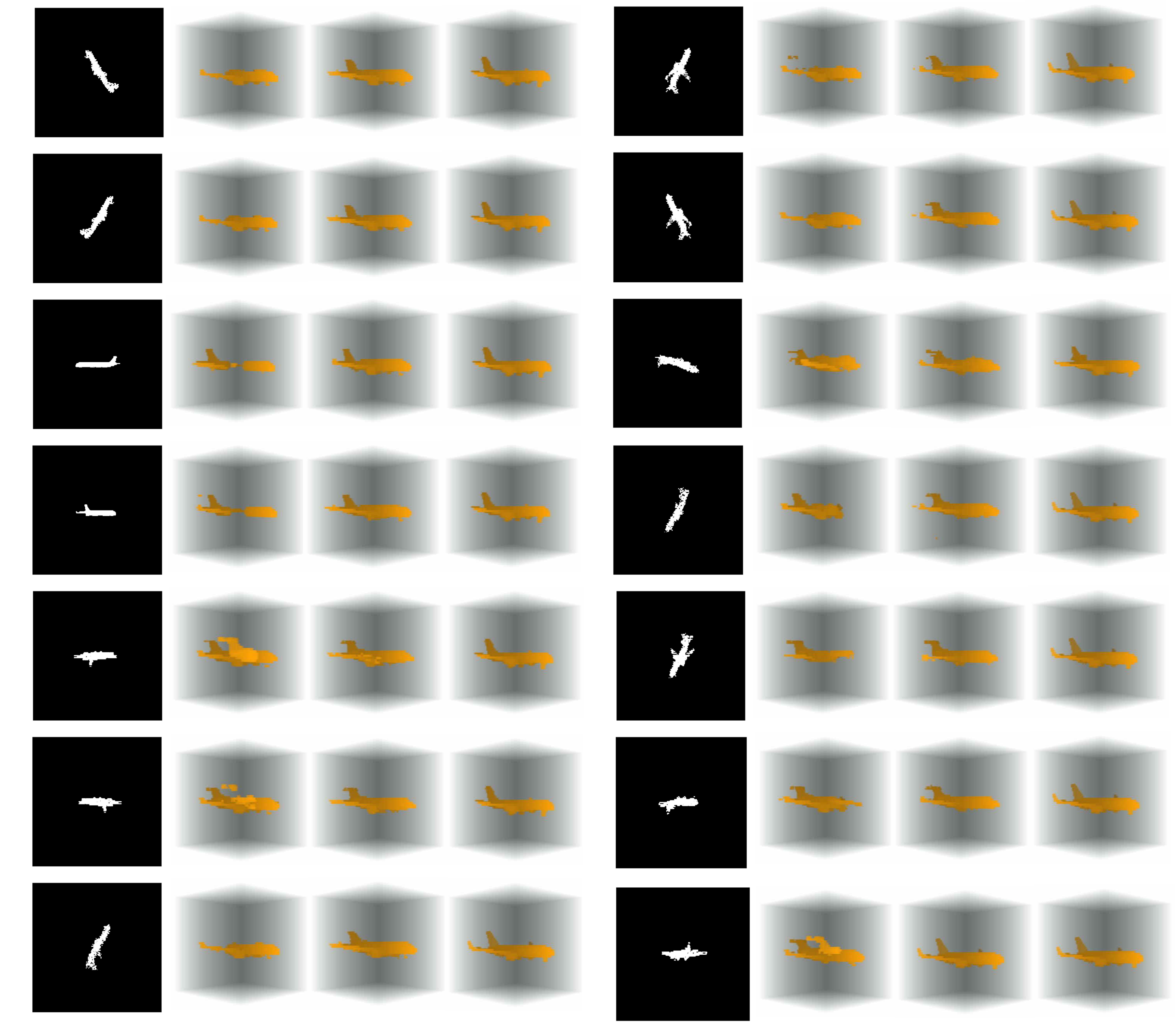}
\caption{Reconstruction visualization from a easy single-view silhouette image for planes(Experiment1). The first column represents picked easy single-view silhouette images of an object. The second column represents the rebuilt 3D shape reconstructed from a single network given the corresponding silhouette images. The third column represents the rebuilt 3D shape reconstructed from the proposed stacked networks. The fourth column represents the ground truth shape.}
\label{fig:8}
\end{figure}

Also, as illustrated in Fig\ref{fig:8}, for 3D reconstruction of planes from a single-view silhouette image, and most part of the shape is available in each silhouette image from a single view, the single network is able to rebuild most part of 3D shape of a plane and some parts are missing such as the tail, the wing, the head, the back. Also, it sometimes fix the tail at the wrong position. However, the proposed stacked networks and its training strategy help the reconstruction to solve these issues and rebuild good shapes. Also, the number of easy single-view silhouette images of planes for test is 3520.

\subsection{Evaluation on Sphere-surface Randomly Located Single-view Silhouette Images (Experiment2)}
For this test, we test 22 CAD models for each category and 180 single-view silhouette images are rendered for the test of each model. A single view for each silhouette image is picked following a different rule Eq.(\ref{eq:9}) from the training to ensure the all the views are randomly located on a sphere around a object and these views are different with the trained views. Where \(l_{1}\in\{1,2,...,18\},l_{2}\in\{1,2,...,18\},\gamma\in(0,1)\) is a random number.

\begin{equation} \label{eq:9}
\alpha_{1}=10(l_{1}-\gamma),\beta_{2}=20(l_{2}-\gamma)
\end{equation}

This experiment is more challenging than experiment1 as that the silhouette images are projected from randomly located single views on the sphere, and the views are different from the views used for training. And we test \(22\times180\) single-view silhouette images in total for each object category.

\begin{figure}
\centering
\includegraphics[width=1.0\linewidth,natwidth=305,natheight=321]{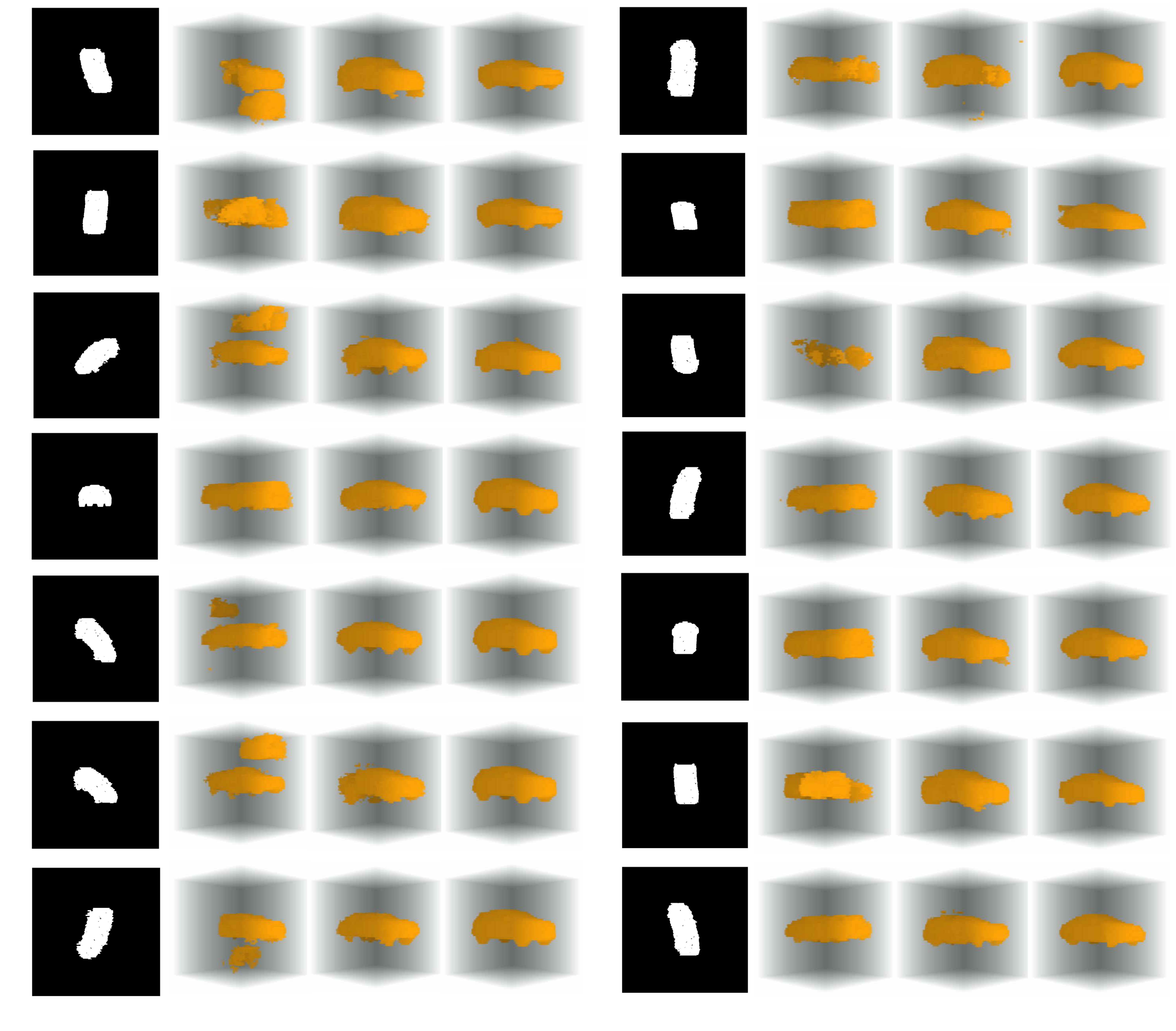}
\caption{Reconstruction visualization from a randomly located single-view silhouette image for cars(Experiment2). The first column represents picked single-view silhouette images of an object. The second column represents the rebuilt 3D shape reconstructed from a single network given the corresponding silhouette images. The third column represents the rebuilt 3D shape reconstructed from the proposed stacked networks. The fourth column represents the ground truth shape.}
\label{fig:9}
\end{figure}

As illustrated in Fig\ref{fig:9}, for reconstruction of cars from a single-view silhouette images when the views are towards randomly located. and for a single view that small part of shape is available or a large part of shape is available, the single network is able to reconstruct a small part of the car, rebuild a coarse 3D shape that is far from a car structure. Also the single network sometimes fails to produce a reasonable shape and much noise occurs in the grid space. However, the stacked networks are able to reconstruct good quality of shapes without those drawbacks. 

\begin{figure}
\centering
\includegraphics[width=1.0\linewidth,natwidth=305,natheight=321]{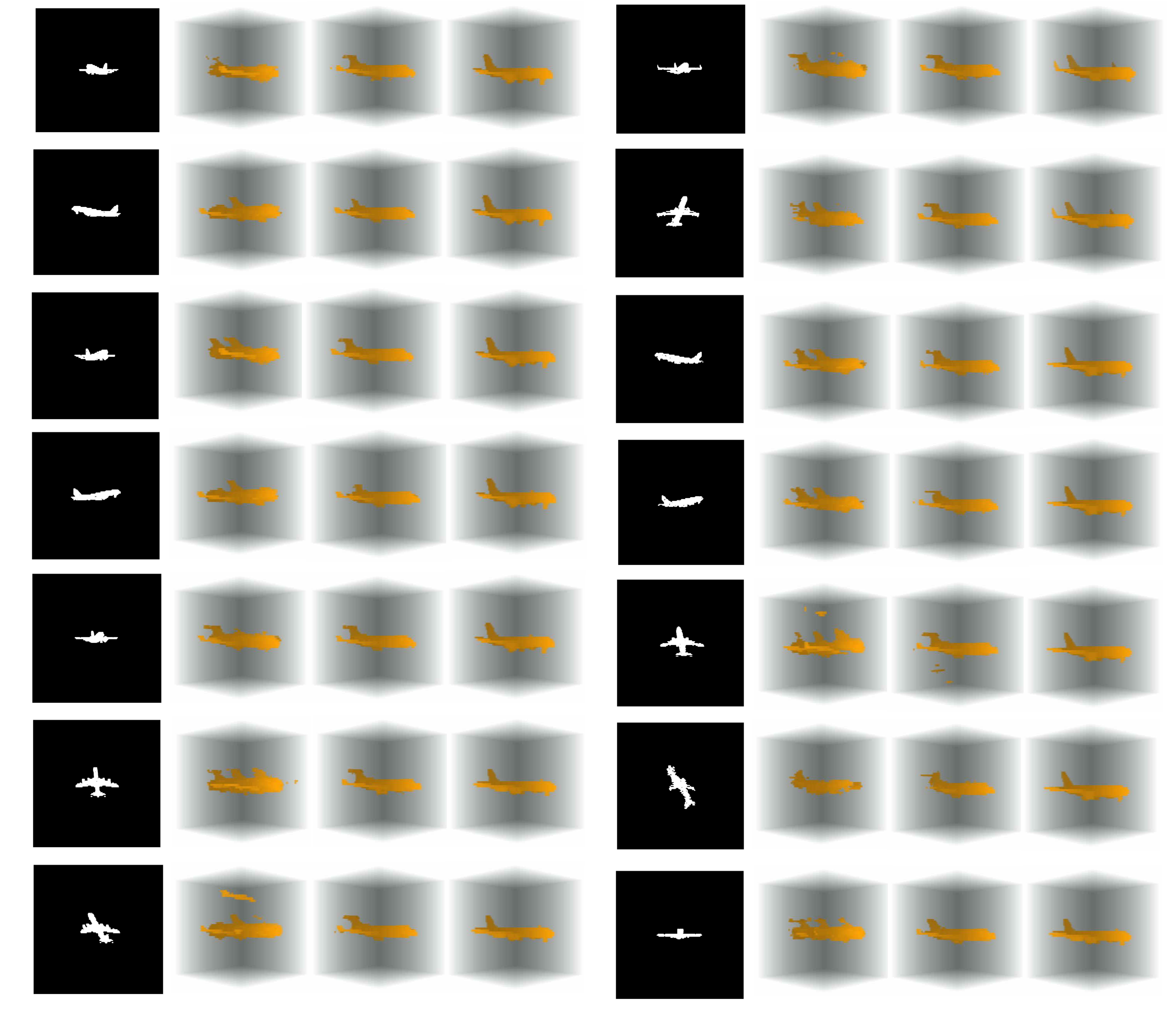}
\caption{Reconstruction visualization from a randomly located single-view silhouette images for planes(Experiment2). The first column represents picked single-view silhouette images of an object. The second column represents the rebuilt 3D shape reconstructed from a single network given the corresponding silhouette images. The third column represents the rebuilt 3D shape reconstructed from the proposed stacked networks. The fourth column represents the ground truth shape.}
\label{fig:10}
\end{figure}

Similarly, as illustrated in Fig\ref{fig:10}, for reconstruction of planes from a single-view reconstruction when the views are randomly located on a sphere surface, and for a single-view silhouette image from views where both a small part of shape or a large part of shape is available, the single network rebuild planes that parts of the plane such as wings, tail are missing or these parts are fixed at the wrong position. Also, it sometimes fails to reconstruct reasonable shapes of planes. However, the proposed stacked networks reconstruct good quality of shapes which overcomes these issues.

\begin{figure}
\centering
\includegraphics[width=1.0\linewidth,natwidth=305,natheight=321]{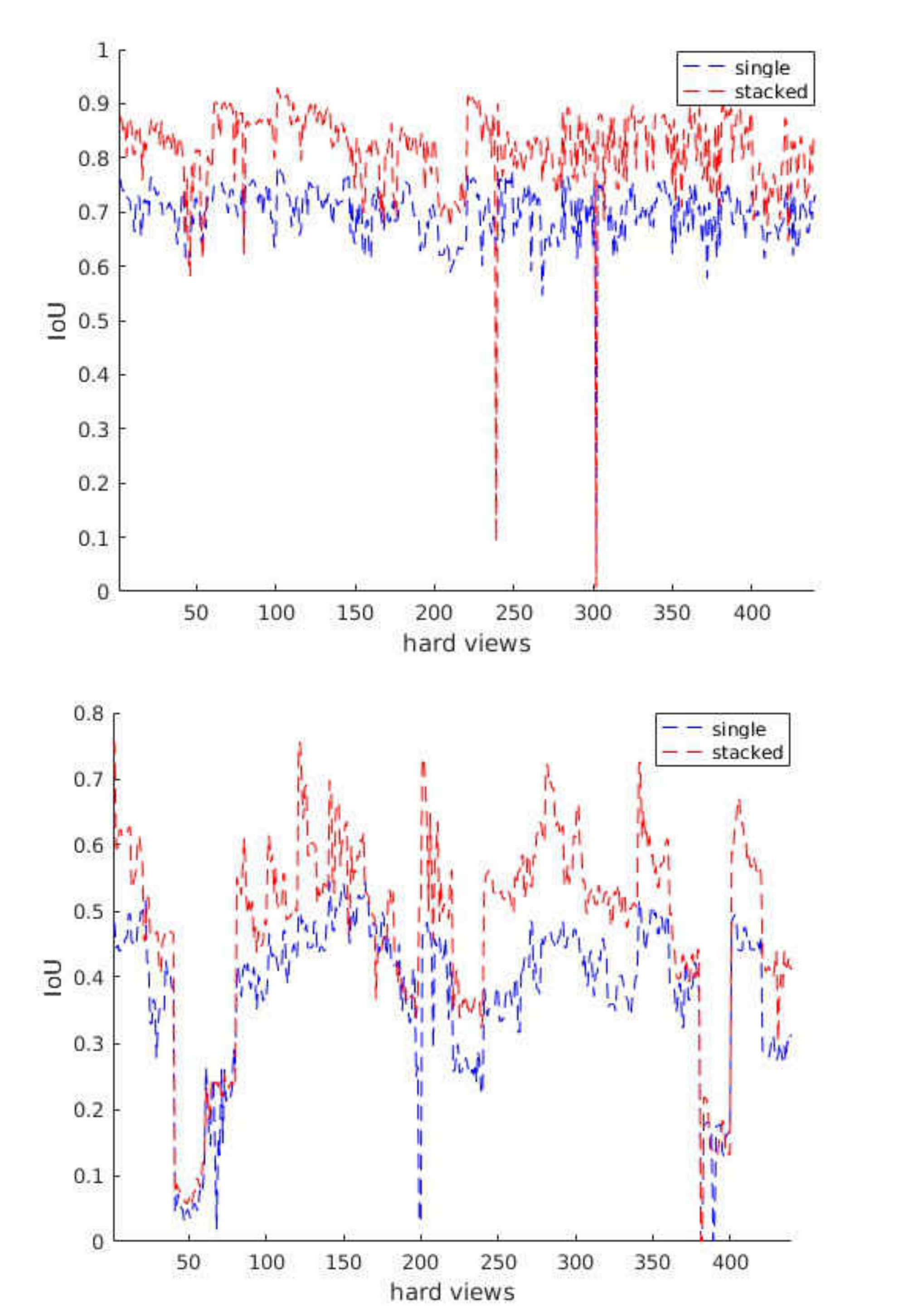}
\caption{Voxel IoU measure of a single network and stacked networks for experiment1. For cars, voxel IoU for each hard single-view silhouette image are represented(top). Similarly, the IoU for hard single-view silhouette of planes are represented(bottom).}
\label{fig:11}
\end{figure}

\begin{table*}
\begin{center}
\begin{tabular}{|l|c|c|c|c|c|c|c|c|}
\hline
 & 3D-R2N2 & S-Hard-E1 & S-All-E1 & SS-Hard-E1 & SS-All-E1 & S-All-E2 & SS-All-E2\\
\hline\hline
Cars & 0.798 & 0.699 & 0.765 & 0.817 & \textbf{0.828} & 0.601 & 0.686\\
Planes & \textbf{0.513} & 0.373 & 0.469 & \textbf{0.473} & \textbf{0.474} & 0.384 & 0.430\\
\hline
\end{tabular}
\end{center}
\caption{Average IoU for car category and plane category. In the table, S represents the single network, SS represents the proposed stacked networks, E1 and E2 represents experiment1 and experiment2, Hard represents hard single-view silhouette images, All represents silhouette images for all test single views.}
\label{tab:1}
\end{table*}

\subsection{Comparison between Stacked Networks,3D-R2N2,Single Networks.}
Comparison between the single network \cite{DiECCV2016}, proposed stacked networks and the state of art 3D-R2N2 \cite{choy20163d} is made. Reconstruction results are calculated in voxel intersection-over-Union(IoU) for the above two experiments. Experiments are conducted with the same configuration for the single network and stacked networks expect 3D-R2N2 which works for the input with RGB textures. As 3D-R2N2 works for RGB images rendered from Shapenet \cite{choy20163d} as input in a specific sequence which dose not works for a single-view silhouette image. We compare IoU of the experiments from both the single network and the stacked networks with the same object category test \cite{choy20163d} in Shapenet. And also compare the average IoU for the same object category from Shapenet that 3D-R2N2 gets from a single-view RGB image \cite{choy20163d}. 

As illustrated in Tab.\ref{tab:1}, the average IoU value of the stacked networks is higher than 3D-R2N2 for cars in the experiment1. Although the stacked networks achieves the same IoU lever for planes that is \(0.04\) lower than \cite{choy20163d}, but we consider our proposed framework is able to build object category specific reconstruction from a single view silhouette image in a good quality also as represented in Fig\ref{fig:6}, Fig\ref{fig:7}, Fig\ref{fig:8}, Fig\ref{fig:9} and Fig\ref{fig:10}. And the network is demonstrated to be able to do 3D reconstruction from single-view with reduced dependence from RGB texture images to 2D silhouette images. Furthermore, the difficulty of experiment in our work is higher than 3D-R2N2 \cite{choy20163d} that the input is reduced to a single-view binary silhouette image. And the views get more variation that the rendered RGB images are projected randomly 5 times on a circle \cite{choy20163d}, while the single-view silhouette images are rendered 180 times from widely located views on a sphere which is towards random single-view configuration. 

Also, the stacked networks achieve higher IoU value than single network Tab.\ref{tab:1}. Even the increased IoU is about 0.1 and 0.07 higher, but as the Fig\ref{fig:6}, Fig\ref{fig:7}, Fig\ref{fig:8}, Fig\ref{fig:9} and Fig\ref{fig:10} demonstrate, this improvement demonstrates that the proposed stacked network reconstruct much higher quality shapes than the single network. In detail, we also demonstrate each IoU for the hard single-view silhouette image in experiment1 between the single network and the stacked networks Fig\ref{fig:11}. The stacked networks achieves distinctly higher IoU than the single network for each single-view silhouette image.

As illustrated in Tab.\ref{tab:1}, for the experiment of randomly located single-view silhouette images, \(74.8737\%\) of 3960 single-view silhouette images of cars achieves high IoU, the percentage for planes is \(52.8283\%\) of 3960 randomly located single-view silhouette images of planes. The IoU value is about \(0.07\) lower than 3D-R2N2 \cite{choy20163d}. We consider contribution as the reconstructed shape is in good quality as illustrated in Fig\ref{fig:9} and Fig\ref{fig:10}. Also, the proposed networks work from single-view silhouette images without RGB textures and the views are widely and randomly located around a sphere in spite of a circle \cite{choy20163d}. Furthermore, in this experiment, all the views are different in the training. 

Also, in order to ensure that the all the average IoU values are credible, we also calculate the standard error, and the standard error is around \(10^{-4}\) which is very low.

\section{Conclusion}

Firstly, the proposed reconstruction framework is represented to reconstruct good quality 3D category specific shape from a single-view silhouette image, while the state of art of 3D reconstruction from a single view depends on RGB textures and other extra input. To our best knowledge, for object specific category, we firstly demonstrate that networks help the task, 3D reconstruction from single view, reduce the dependence of RGB textures to binary segmentation.  

Secondly, a novel network design is proposed for stacked hierarchical networks combined of a number of single networks. Also an end-to-end training strategy for the stacked networks are proposed to help the complex stacked networks find global optimization solution during training. As the application of the 3D reconstruction from a single-view silhouette image demonstrates, this contribution improves the networks' ability in this reconstruction task to overcome many drawbacks and even failures. Further work to the development of complex stacked networks in this end-to-end training direction may be promising. 

Thirdly, the proposed networks are demonstrated to build 3D reconstruction from a single view that are both widely and randomly located on a sphere around a object. While to our best knowledge, other network frameworks are demonstrated to build 3D reconstruction from a single view that is randomly located on a circle or partially located on a sphere. Therefore, the proposed network framework is able to reconstruct 3D shape more towards a randomly located single view than the state of art. Further work may include 3D reconstruction for a single view randomly located in space in spite of the surface of a sphere.

Finally, although the stacked hierarchical networks are trained end to end, but the replication layer is not currently in the end to end training strategy. Because we currently reconstruct the ground truth 3D shape to ensure that the 3D shape is fixed without rotation and translation. The future work includes the development of a more completed training strategy for the whole shape reconstruction framework, and some new single networks may be added in the proposed framework for the rotation and translation of a reconstructed shape. 
{\small
\bibliographystyle{ieee}
\bibliography{egbib}
}

\end{document}